\documentclass[runningheads]{llncs}

\usepackage{eccv}

\usepackage{eccvabbrv}

\usepackage{graphicx}
\usepackage{booktabs}

\usepackage[accsupp]{axessibility}  

\usepackage[pagebackref,breaklinks,colorlinks,citecolor=eccvblue]{hyperref}

\usepackage{orcidlink}
\usepackage{multirow}
\usepackage{bbding}
\usepackage{adjust box}

\usepackage{rotating}

\begin{document}

\title{CutDiffusion: A Simple, Fast, Cheap, and Strong Diffusion Extrapolation Method}

\titlerunning{CutDiffusion}

%


\author{Mingbao Lin\inst{1} \and
Zhihang Lin\inst{2} \and
Wengyi Zhan\inst{2}  \and
Liujuan Cao\inst{2}\thanks{Correspondence: \email{caoliujuan@xmu.edu.cn}; First Author: \email{linmb001@outlook.com}}  \and
Rongrong Ji\inst{2}}

\authorrunning{Lin et al.}


\institute{$^1$Skywork AI \qquad
$^2$Xiamen University\\
\url{https://github.com/lmbxmu/CutDiffusion}}

\maketitle

\begin{abstract}

Transforming large pre-trained low-resolution diffusion models to cater to higher-resolution demands, \emph{i.e.}, diffusion extrapolation, significantly improves diffusion adaptability.
We propose tuning-free CutDiffusion, aimed at simplifying and accelerating the diffusion extrapolation process, making it more affordable and improving performance.
CutDiffusion abides by the existing patch-wise extrapolation but cuts a standard patch diffusion process into an initial phase focused on comprehensive structure denoising and a subsequent phase dedicated to specific detail refinement.
Comprehensive experiments highlight the numerous almighty advantages of CutDiffusion:
(1) simple method construction that enables a concise higher-resolution diffusion process without third-party engagement;
(2) fast inference speed achieved through a single-step higher-resolution diffusion process, and fewer inference patches required;
(3) cheap GPU cost resulting from patch-wise inference and fewer patches during the comprehensive structure denoising;
(4) strong generation performance, stemming from the emphasis on specific detail refinement.

\end{abstract}

\section{Introduction}
\label{sec:intro}

As a member of diffusion architecture models~\cite{ho2020ddpm,song2020ddim}, also called a latent diffusion model~\cite{rombach2022SD}, stable diffusion, primarily used to generate detailed images conditioned on text descriptions, has been considered to be a crucial piece of the ongoing bloom of generative artificial intelligence.
Coupled with its realistic image generation, stable diffusion is open-sourced and supports running on most consumer hardware equipped with a modest 4 GB VRAM GPU. Thus, it marked a departure from many proprietary text-to-image models such as DALL-E series~\cite{ramesh2021zero,ramesh2022hierarchical,betker2023improving} and Midjourney program~\cite{midjourney}, accessible only via cloud services.

One of the dominant real-world applications is to generate high-resolution images.
Initial releases of stable diffusion were trained on a 512 $\times$ 512 resolution dataset, resulting in quality issues like degradation and inaccuracies when user-specified resolution deviates from its ``expectation''~\cite{dhariwal2021ldm}.
Though the subsequent version 2.0 strengthened the ability to generate 768 $\times$ 768 resolution images~\cite{dhariwal2021ldm} and Stable Diffusion XL (SDXL) released native 1024 $\times$ 1024 resolution~\cite{podell2023sdxl}, retraining of stable diffusion models is slow because of its repeated, sequential nature, and eats a lot of memory because of the operations in latent pixel space, which becomes huge for high-resolution images.
It was reported to train the initial release using 256 Nvidia A100 GPUs on Amazon Web Services for a total of 150,000 GPU-hours, at a cost estimation of \$600,000~\cite{stable-diffusion-1}.
The charge drastically increases in higher resolution, which as consequences is almost prohibitive for individuals, even for most small- and medium-sized enterprises, to train large diffusion models one by one for a certain resolution size, in particular, taking into account the ultra-high resolution such as 4096 $\times$ 4096 or 8192 $\times$ 8192.

\begin{table}[!tb]
    \caption{\label{tab:cons-and-pros}Comparison for existing diffusion extrapolation methods.}
    \vspace{-1.0em}
    \centering
    \setlength\tabcolsep{6pt}
    \resizebox{\textwidth}{!}{
    \begin{tabular}{c|cccccc}
    \toprule
         \multirow{2}{*}{Method} 
         & Tuning  & Fast & Cheap & Equal & \multirow{2}{*}{Performance} 
         & \multirow{2}{*}{Simplicity} \\
         & Free & Infer   & GPU  
         & Params &       &      \\
         
    \midrule
         DiffFit~\cite{xie2023diff-fit}                     
         & \XSolidBrush    & \Checkmark      &\XSolidBrush   & \XSolidBrush   & \Checkmark    & \XSolidBrush \\
        
         
         Self-Cascade~\cite{guo2024make-a-cheap-scaling}    
         & \XSolidBrush    & \Checkmark      &\XSolidBrush   & \XSolidBrush  & \Checkmark    & \XSolidBrush  \\
         
         SDXL-DI~\cite{podell2023sdxl}                      
         & \Checkmark        & \Checkmark      &\XSolidBrush   &\Checkmark   & \XSolidBrush  & \Checkmark  \\
         
         Attn-SF~\cite{jin2023logn}                         
         & \Checkmark      & \Checkmark      &\XSolidBrush   & \Checkmark   
         & \XSolidBrush  & \Checkmark  \\
         
         ScaleCrafter~\cite{he2023scalecrafter}             
         & \Checkmark      & \XSolidBrush    &\XSolidBrush   & \Checkmark   
         &\XSolidBrush  & \XSolidBrush  \\ 
         
         MultiDiffusion~\cite{bar2023multidiffusion}        
         & \Checkmark       & \XSolidBrush 
         &\Checkmark        &\Checkmark   
         & \XSolidBrush     & \Checkmark   \\
         
         DemoFusion~\cite{du2023demofusion}                 
         & \Checkmark      & \XSolidBrush    &\Checkmark     & \Checkmark   
       & \Checkmark & \XSolidBrush \\
         
         CutDiffusion (Ours)                                
         & \Checkmark       & \Checkmark      & \Checkmark    & \Checkmark  
         & \Checkmark    & \Checkmark  \\
         
    \bottomrule
    \end{tabular}
    }
\vspace{-2.0em}
\end{table}

To address the impediments of training an initial model, diffusion extrapolation, which relies on a well-pretrained low-resolution model and achieves higher-resolution image diffusion, has gained significant interest.
Recent research workers opt for fine-tuning involvement~\cite{xie2023diff-fit,guo2024make-a-cheap-scaling}, or even tuning-free implementation~\cite{bar2023multidiffusion,du2023demofusion,jin2023logn,he2023scalecrafter}, to realize more user-specific generation outputs.
In Table\,\ref{tab:cons-and-pros}, we assess major pros and cons of these methods considering factors of tuning-free, fast inference speed, cheap GPU costs, equal parameters, performance and method simplicity.
We embark on method analyses here and specific reports will be given in Sec.\,\ref{sec:experimentation}.
On the lines of fine-tuning involvement, most methods make efforts for a parameter-efficient optimization to enable fast adaptation to ``unexpected'' resolution generation. For example,
DiffFit~\cite{xie2023diff-fit} fine-tuned the bias term and newly-added scaling factors in specific layers and Self-Cascade~\cite{guo2024make-a-cheap-scaling} upgraded time-aware feature upsampling modules to inject the side information from newly acquired higher-resolution data.
%
%
%
These tuning methods excel in the accurate portrayal of local structural details. Their ability to scale to the target resolution in one step facilitates rapid inference speeds.
However, direct inference at higher resolutions incurs sustainable peak GPU memory expenses due to more pixels, and parameter tuning adds complexity, sometimes extra parameters~\cite{xie2023diff-fit,guo2024make-a-cheap-scaling}.

In tuning-free implementation, all methods keep the same parameter amount as the original models. A straightforward approach is direct higher-resolution inference, which offers fast speed but requires high GPU consumption, similar to fine-tuning methods.
Results in~\cite{du2023demofusion} with SDXL (denoted as SDXL-DI) showed poor object repetition, twisted object and unreasonable structure, \emph{etc}.
Attn-SF~\cite{jin2023logn} simplified its methodology by scaling attention entropy in balance with the token quantity, which however still endures poor quality of object repetition.
ScaleCrafter~\cite{he2023scalecrafter} identified the limited receptive field of convolutional kernels, and emphasized on complex adjustment of dilated stride, injected step, \emph{etc}. Also, larger kernels affect inference speed.
Another tuning-free implementation performs diffusion extrapolation in a patch-wise manner, reducing the inference GPU costs. For instance, MultiDiffusion~\cite{bar2023multidiffusion} split an entire higher-resolution noise into multiple overlapping patches and binded them together through multiple diffusion generation processes.
Albeit its simplicity merit, lack of global information leads to issues like repetitive and distorted generation.
Recent DemoFusion~\cite{du2023demofusion} constructed residual connection and dilated sampling to incorporate global semantic information. Despite its better performance, these operations make DemoFusion sophisticated.
Moreover, DemoFusion adopts a fashion of progressive upscaling, where for example, the generation of a 4096 $\times$ 4096 image is constructed on intermediate diffusions of 1024 $\times$ 1024, 2048 $\times$ 2048, and 3072 $\times$ 3072, causing more diffusion steps and slower inference latency.
%
%
%

We dedicate our efforts to exploring the potential of tuning-free diffusion extrapolation through a patch-wise generation approach. Our primary objective is to simplify and expedite the diffusion extrapolation process, ultimately making it more cost-effective and enhancing its overall performance.
With this goal in mind, we present CutDiffusion in this paper, an almighty method that cuts the conventional patch diffusion process into two separate stages, each incorporating distinct sub-patch partitioning methodologies.
%
%
During the preliminary stage, non-overlapping patches are arbitrarily sampled from an independent Gaussian distribution and recovered in a spatial pixel interaction fashion, ensuring that each patch focuses on denoising the entire structure. In essence, every individual patch strives to comply with the same textual directive and produce analogous content.
In continuation of the subsequent stage, we reassemble the identical spatial pixels of the first-phase sub-patches to create a more refined higher-resolution Gaussian noise. 
Then, we follow MultiDiffusion~\cite{bar2023multidiffusion} to develop overlapping patches, where each individual patch is designed to focus on refining and enhancing partial details within the overall image.
As listed in Table\,\ref{tab:cons-and-pros}, in addition to its tuning-free completion and absence of parameter introduction, the proposed CutDiffusion also offers substantial advantages in terms of:

(1) Simple method construction. CutDiffusion is constructed upon a standard diffusion process, only altering sub-patch sampling, which eliminates complex third-party engagement and creates a more concise higher-resolution process.

%

%
(2) Fast inference speed. CutDiffusion performs a single-step upscaling to the target resolution, an efficient alternative to progressive upscaling, and necessitates fewer inference patches during denoising the comprehensive structure.

%

(3) Cheap GPU cost. CutDiffusion employs a patch-wise approach for extrapolation and decreases the quantity of non-overlapping patches required for structure denoising, effectively mitigating the necessity for peak GPU memory.


%
(4) Strong generation performance. Albeit its single-step upscaling approach and absence of third-party engagement, CutDiffusion gracefully executes partial detail refinement, consequently achieving superior generation performance.

\section{Related Work}

\subsection{Diffusion Models}

Diffusion models, often known as probabilistic diffusion models~\cite{ho2020ddpm}, fall under latent variable generative models. 
A classic diffusion model is comprised of the forward process, the reverse process, and the sampling methodology~\cite{chang2023design}.
The forward process initiates at a given starting point, and then consistently introduces noise into the system.
The reverse process generally entails teaching a neural network to remove Gaussian noise from blurred images. The model is designed to counteract the noise addition process in an image, and employed for image generation by refining and denoising a randomly noisy image.
The sampling mechanism operates by navigating parameters and diminishing the variance of the samples, resulting in more accurate approximations of the desired distribution.
DDIM~\cite{song2020ddim} allows the utilization of any model trained on DDPM loss~\cite{ho2020ddpm} to sample with a few steps omitted, trading a modifiable quantity of quality. 
LDM~\cite{rombach2022SD} initially compresses images, applies a diffusion model to represent the distribution of encoded images, and finally decodes them into images.
The supreme performance has made it widely applied to generate text-driven images.

\subsection{Diffusion Extrapolation} 


\subsubsection{Fine-tuning Involvement.}
It is a logical idea to fine-tune pre-trained diffusion models to accommodate the generation of higher-resolution images. 
Owing to the considerable training costs of these models, seeking parameter-efficient optimization methods has become a central focus for researchers and practitioners in the fine-tuning field.
In their research, Xie \emph{et al}.~\cite{xie2023diff-fit} took an innovative approach by freezing most of the parameters within the latent diffusion model. They focused on training specific parameters such as the bias term, normalization, and the class conditioning module. Additionally, they incorporated learnable scale factors into the model to further enhance its performance and adaptability.
Guo \emph{et al}.~\cite{guo2024make-a-cheap-scaling} developed an innovative cascade framework that included a low-resolution model at its initial scale and additional new modules at a higher scale. 

\subsubsection{Tuning-free Implementation.}
Another line of researchers seek for tuning-free diffusion extrapolation. 
%
%
In Attn-SF~\cite{jin2023logn}, lower resolution images showed incomplete objects, while higher resolution ones had repetitive chaos. Therefore, a scaling factor was added to address attention entropy shifts and fix defects.
He \emph{et al}.~\cite{he2023scalecrafter} identified limited kernel perception as causing object repetition and structural issues in diffusion models. They suggested dynamic perception field adjustment and proposed dispersed convolution and noise-damped guidance for ultra-high-resolution generation.
MultiDiffusion~\cite{bar2023multidiffusion} demonstrated the capability of combining several overlapping denoising paths to create panoramic images.
The study DemoFusion~\cite{du2023demofusion} has innovatively enhanced the injection of global information. This enhancement is achieved by using intermediate noise-inversed representations as skip residuals. Additionally, the use of dilated sampling in the denoising paths further contributes to the improvement.

\begin{figure}[!t]
    \centering
    \includegraphics[width=0.95\textwidth]{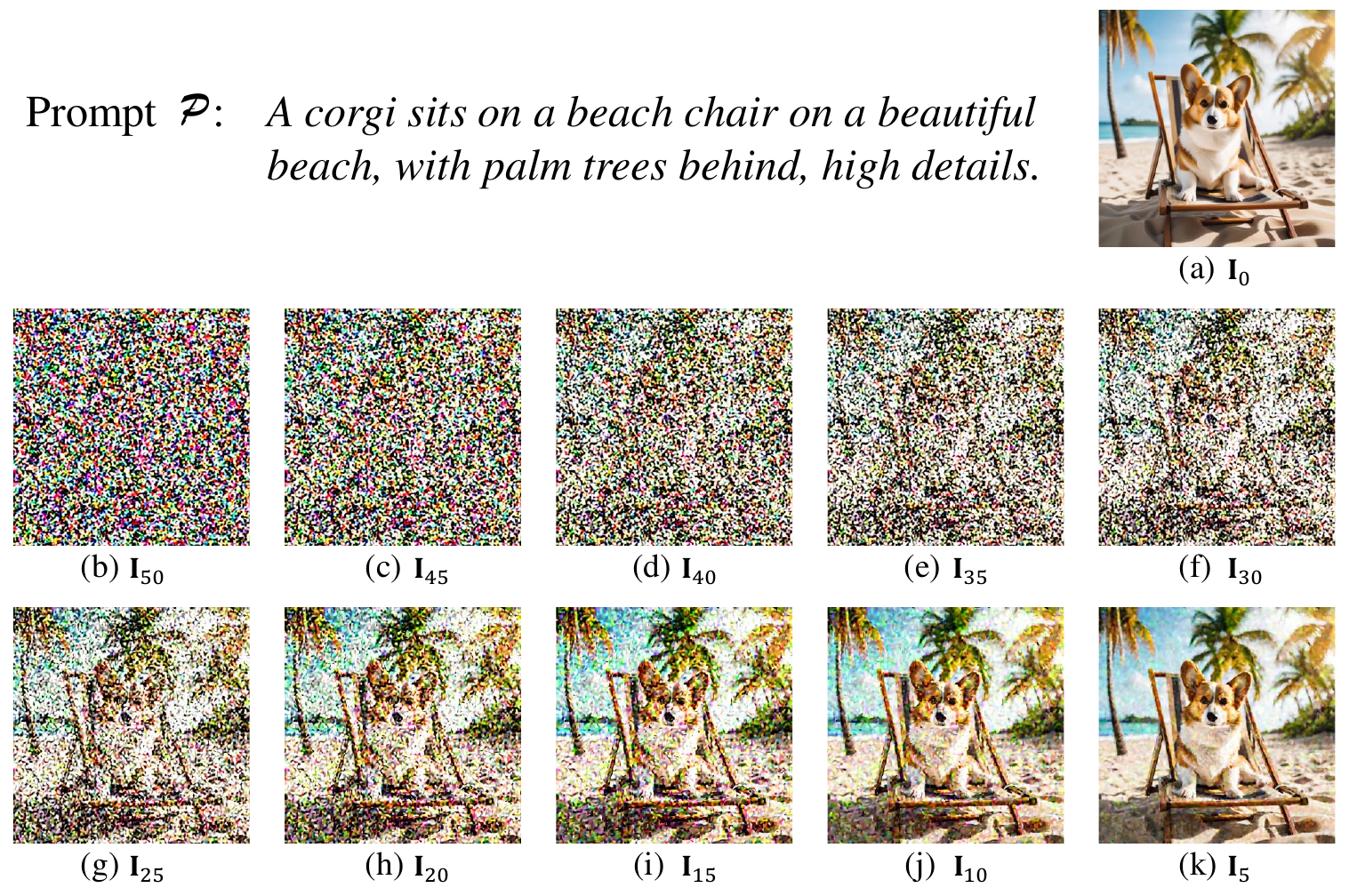}
    \caption{Visualization of decoded $\mathbf{I}_t$. The $\mathbf{I}_0$ denotes final clean image (1024$\times$1024).}
    \label{fig:visualization}
    \vspace{-1.0em}
\end{figure}

\section{Methodology}

\subsection{Background}\label{sec:backgrounds}

\textbf{Latent Diffusion Model.}
Let $\mathbf{z}_T \in \mathbb{R}^{h \times w \times c}$ represent a noise tensor following a Gaussian distribution in the latent space. A large pre-trained latent diffusion model (LDM)~\cite{rombach2022SD} utilizes its three components to denoise $\mathbf{z}_T$ conditioned on a user-specified text prompt $\mathcal{P}$, resulting in a clean image $\mathbf{I}_0 \in \mathbb{R}^{H \times W \times 3}$:

(1) A text encoder $\tau_{\theta}$, transforms conditional text prompt $\mathcal{P}$ to an embedding space $\tau_{\theta}(\mathcal{P})$. This transformation allows for a more comprehensive and meaningful representation of the user-specified text prompt.

(2) A U-Net $\varepsilon_{\theta}$, executes consecutive $T$ denoising diffusion steps, each of which estimates the noise and recovers a cleaner latent $\mathbf{z}_{t-1}$ from $\mathbf{z}_t$:
\begin{equation}
\label{eq:denoising}
    \mathbf{z}_{t-1} = \sqrt{\frac{\alpha_{t-1}}{\alpha_t}}\mathbf{z}_t + \Bigg(\sqrt{\frac{1}{\alpha_{t-1}} - 1} - \sqrt{\frac{1}{\alpha_t} - 1} \Bigg)\cdot \varepsilon_{\theta}\big(\mathbf{z}_t,t,\tau_{\theta}(\mathcal{P})\big),
\end{equation}
where $t = T \rightarrow 1$ and $\{\alpha_t\}_{t=1}^T$ is a set of prescribed variance schedules.

(3) A VAE decoder $\mathcal{D}$, is responsible for decoding the final image. It accomplishes this by converting the latent representation back into pixel space $\mathbf{I}_0 = \mathcal{D}(\mathbf{z}_0)$.
We visualize the decoding of latent $\mathbf{z}_t$, \emph{i.e.}, $\mathbf{I}_t = \mathcal{D}(\mathbf{z}_t)$ and the final clean image $\mathbf{I}_0$ in Fig.\,\ref{fig:visualization}, where our motivation originates as described in Sec.\,\ref{sec:frameowork}.

The above simply introduces essential notations for subsequent content. We recommend the original paper~\cite{rombach2022SD} for a thorough understanding of LDM.

\begin{figure}[!t]
    \centering
    \includegraphics[width=0.95\linewidth]{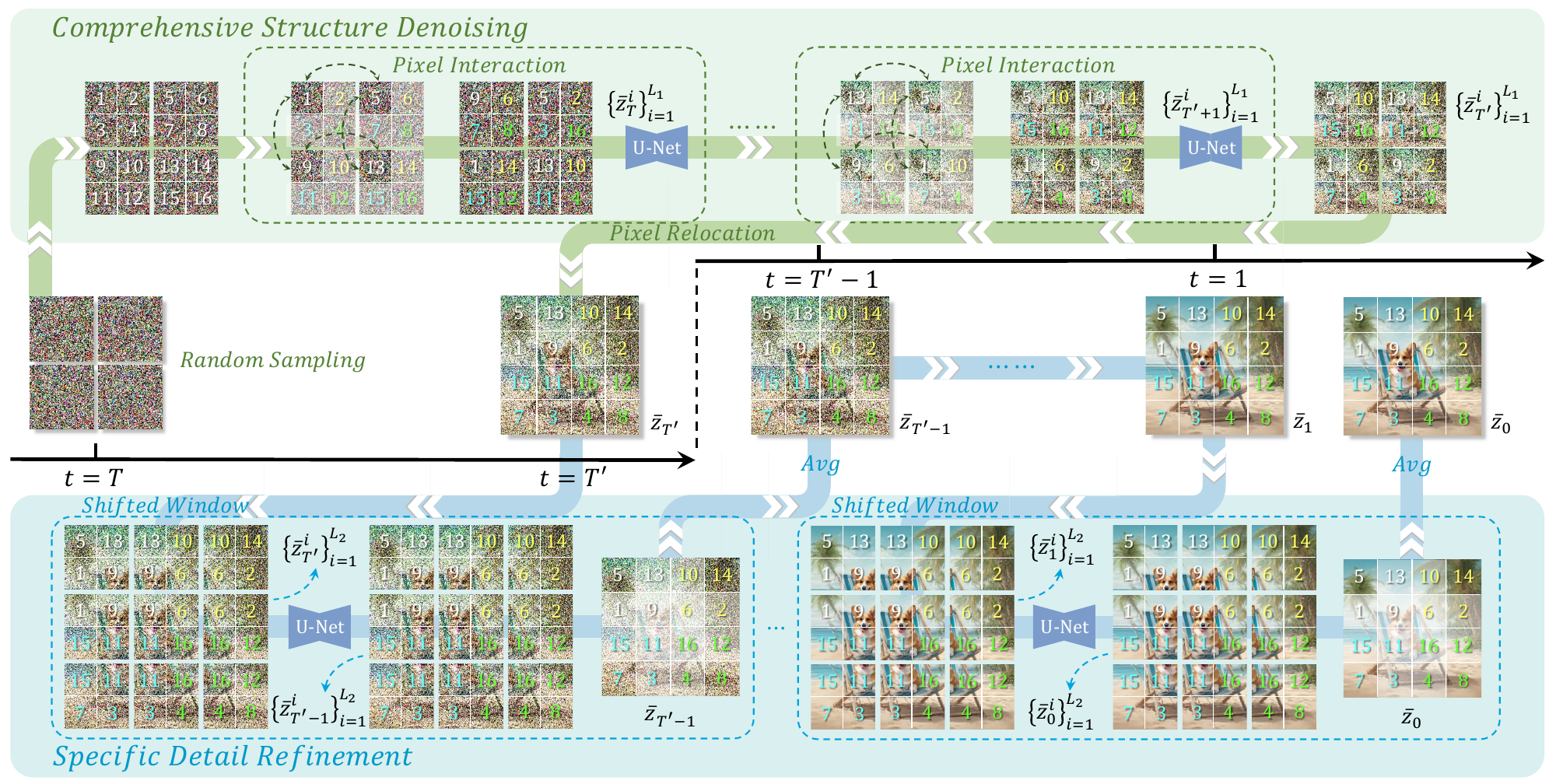}
    \caption{
    CutDiffusion framework where ``$\mathbf{>>}$'' denotes workflow. Comprehensive structure denoising assigns similar content to each non-overlapping patch. Specific detail refinement enhances details within overlapping patches. Best view with zooming in.}
    \label{fig:framework}
    \vspace{-1.0em}
\end{figure}

\subsubsection{Diffusion Extrapolation.}

An LDM model is mostly pre-trained on a fixed resolution size, such as $H=1024$, $W=1024$ for SDXL~\cite{podell2023sdxl}, leading to latent resolution $h = 128$, $w=128$. Diffusion extrapolation refers to the case of generated resolution $\bar{h}$ ($\bar{H}$), $\bar{w}$ ($\bar{W}$) beyond the pre-trained resolution.
A variety of existing methods, as referenced in studies like~\cite{xie2023diff-fit, guo2024make-a-cheap-scaling, jin2023logn, he2023scalecrafter}, engage in the process of conducting direct higher-resolution inference, which consequently contributes to a significant challenge in terms of sustainable peak GPU expenses.
Recent patch-wise techniques~\cite{bar2023multidiffusion,du2023demofusion} utilized a shifted window to sample overlapping patches $\{\bar{\mathbf{z}}_T^i\}_{i=1}^{L}$ from higher-resolution latent $\bar{\mathbf{z}}_T \in \mathbb{R}^{\bar{h} \times \bar{w} \times c}$, where $\bar{\mathbf{z}}_T^i \in \mathbb{R}^{h\times w\times c}$ and $L = (\frac{\bar{h}-h}{d_h}+1)\cdot(\frac{\bar{w}-w}{d_w}+1)$ is patch number, $d_h$ and $d_w$ are vertical and horizontal strides.
Then, patch-wise denoising via Eq.\,(\ref{eq:denoising}) is conducted to derive $\bar{\mathbf{z}}_{t-1}$ where the overlapping sections are calculated by taking the average. Finally, $\bar{\mathbf{z}}_{0}$ is decoded to a higher-resolution image $\bar{\mathbf{I}}_0 \in \mathbb{R}^{\bar{H} \times \bar{W} \times 3}$.

Most of these methods require complex third-party engagement such as fine-tuning~\cite{xie2023diff-fit,guo2024make-a-cheap-scaling}, low-resolution image generation~\cite{du2023demofusion}, progressive upscaling~\cite{du2023demofusion}, \emph{etc.}
We develop our method by employing a patch-based strategy, which is specifically designed to simplify and effectively manage the complex operations involved in these intricate processes, ensuring a more efficient and user-friendly approach.

\subsection{Motivation and Framework}\label{sec:frameowork}
%
%
%


\subsubsection{Motivation.}
In Fig.\,\ref{fig:visualization}, LDM progressively denoises random noise in Fig.\,\ref{fig:visualization}(b) via iterative steps until a predetermined count ($t = 50 \rightarrow 1$) is reached. The resulting Fig.\,\ref{fig:visualization}(a) captures the prompt $\mathcal{P}$ concept.
Upon examining the progression from Fig.\,\ref{fig:visualization}(b) to Fig.\,\ref{fig:visualization}(k), and ultimately the clean image in Fig.\,\ref{fig:visualization}(a), we can deduce that the diffusion process involves an initial stage of denoising the entire structure, followed by a subsequent stage of refining partial details.
At earlier time steps, various contour profiles become more recognizable. At step $t = 40$, the contour profile resembling a dog becomes clear. As we move to step $t = 35$, the tree contour starts to form. Further to step $t = 30$, the chair's silhouette begins to emerge, making it more distinct, and finally by step $t = 25$, the profiles of the beach, sky, and sea become clearer, allowing for a better overall structure of the scene.
During later time steps from $t = 20$ to $t = 5$, the refinement of local details within the generated contour profiles takes place to enhance the realism of the objects. For instance, the sand appears to collapse around the chair legs, the sky and clouds seamlessly blend together, the dog's fur transforms into a more natural, limp state, and numerous other improvements occur.

Consequently, we recognize a fresh perspective that the diffusion extrapolation can be developed by adhering to the law of progressive image generation. This involves prioritizing comprehensive structure denoising initially, followed by refining specific details, for a more seamless higher-resolution diffusion image.

\subsubsection{Framework.}
To develop concise diffusion extrapolation, we introduce the framework of our CutDiffusion in Fig.\,\ref{fig:framework}. In the premise of a standard diffusion process, CutDiffusion incorporates a patch sampling approach as its only modification.

%
%

In the comprehensive structure denoising phase, we randomly sample multiple non-overlapping patches $\{\bar{\mathbf{z}}_{T}^i\}_{i=1}^{L_1}$ from Gaussian noise, where $\bar{\mathbf{z}}_{T}^i \in \mathbb{R}^{h \times w \times c}$. We then perform reverse diffusion steps $t = T \rightarrow T'$ to denoise each patch's structure. Identical spatial pixels undergo random interactions to enhance textual directive uniformity and generate similar content within denoised patches $\{\bar{\mathbf{z}}_{T'}^i\}_{i=1}^{L_1}$.
In the specific detail refinement stage, we reassemble the structurally enhanced higher-resolution Gaussian noise, $\bar{\mathbf{z}}_{T'} \in \mathbb{R}^{\bar{h} \times \bar{w} \times c}$, by placing identical spatial pixels of $\{\bar{\mathbf{z}}_{T'}^i\}_{i=1}^{L_1}$ adjacently. We then extract overlapping patches $\{\bar{\mathbf{z}}_{T'}^i\}_{i=1}^{L_2}$ from $\bar{\mathbf{z}}_{T'}$ using a shifted window, 
and finally, the reverse diffusion process continues to restore locality for realistic image generation.

%

\subsection{Comprehensive Structure Denoising}
\label{sec:denoising}
%


To generate a higher-resolution image $\bar{\mathbf{I}}_0 \in \mathbb{R}^{\bar{H} \times \bar{W} \times 3}$, one could infer from higher-resolution latent noise $\bar{\mathbf{z}}_T \in \mathbb{R} ^{\bar{h} \times \bar{w} \times c}$~\cite{xie2023diff-fit, guo2024make-a-cheap-scaling, jin2023logn, he2023scalecrafter}, but this leads to significant peak GPU memory usage. Another line employs a shifted window to sample overlapping patches $\{\bar{\mathbf{z}}_T^i\}_{i=1}^{L}$ from $\bar{\mathbf{z}}_T$ where $\bar{\mathbf{z}}^i_L \in \mathbb{R}^{h \times w \times c}$ and perform patch-wise inference~\cite{bar2023multidiffusion,du2023demofusion}. We show our comprehensive structure denoising facilitates the sampling process and reduces the required number of patches.


As detailed in Sec.\,\ref{sec:backgrounds}, the shifted window technique yields $L = (\frac{\bar{h}-h}{d_h}+1)\cdot(\frac{\bar{w}-w}{d_w}+1)$ patches. In contrast, we propose setting $L_1 = \frac{\bar{h} \cdot \bar{w}}{h \cdot w}$, corresponding to the target scale. For example, with $h = w = 128$, $d_h = d_w = 64$, and $\bar{h} = \bar{w} = 384$, the shifted window generates 25 patches, while ours requires only 9 patches. Our method relies on two simple intuitions: (1) Each patch focuses on denoising the entire structure, adhering to the same textual directive and producing similar content, unlike the shifted window that recovers partial details. (2) A minimum of $L_1 = \frac{\bar{h} \cdot \bar{w}}{h \cdot w}$ patches suffices to reassemble a higher-resolution latent $\bar{\mathbf{z}}_T \in \mathbb{R}^{\bar{h} \times \bar{w} \times c}$.

As shown in the top-left of Fig.\,\ref{fig:framework}, we sample non-overlapping patches $\{\bar{\mathbf{z}}_{T}^i\}_{i=1}^{L_1}$ from the Gaussian noise and conduct patch-wise denoising as $t = T \rightarrow T'$:
\begin{equation}
\label{eq:denoising-individual}
    \bar{\mathbf{z}}^i_{t-1} = \sqrt{\frac{\alpha_{t-1}}{\alpha_t}}\bar{\mathbf{z}}^i_t + \Bigg(\sqrt{\frac{1}{\alpha_{t-1}} - 1} - \sqrt{\frac{1}{\alpha_t} - 1} \Bigg)\cdot \varepsilon_{\theta}\big(\bar{\mathbf{z}}^i_t,t,\tau_{\theta}(\mathcal{P})\big).
\end{equation}

\subsubsection{Pixel Interaction.}
%
%
While Eq.\,(\ref{eq:denoising-individual}) generates latents from the same textual description $\mathcal{P}$, it only ensures images related to the same subject, not containing similar content. We therefore introduce spatial pixel interaction among distinct patches. As shown in the top of Fig.\,\ref{fig:framework}, before denoising, we randomly exchange pixels from different patches with the same spatial coordinate ($x, y$) as:
\begin{equation}
    (\bar{\mathbf{z}}_t^1)_{x, y, :}, (\bar{\mathbf{z}}_t^2)_{x, y, :}, ..., (\bar{\mathbf{z}}_t^{L_1})_{x, y, :} = randperm(\{\bar{\mathbf{z}}_{t}^i\}_{i=1}^{L_1}),
\end{equation}
where $randperm()$ returns a random permutation of its input sequence, enabling pixels to contribute to the denoising of other images and promoting similarity in content generation across patches. Sec.\,\ref{sec:ablation} studies the details.

\subsection{Specific Detail Refinement}
\label{sec:refinement}

%
After steps $t = T \rightarrow T'$, comprehensive object structures emerge within low-resolution patches $\{\bar{\mathbf{z}}^i_{T'}\}_{i=1}^{L_1}$. Our approach strategically reassembles pixels from these patches to form a target higher-resolution latent representation. We then refine specific details through the remaining steps, $t = T' \rightarrow 1$, ultimately obtaining a higher-resolution latent $\bar{\mathbf{z}}_0 \in \mathbb{R}^{\bar{h} \times \bar{w} \times c}$ to decode the final clean image.

\subsubsection{Pixel Relocation.}
We previously established the viewpoint that a minimum of $L_1 = \frac{\bar{h} \cdot \bar{w}}{h \cdot w}$ low-resolution patches $\{\bar{\mathbf{z}}^i_{T'}\}_{i=1}^{L_1}$, where $\bar{\mathbf{z}}^i_{T'} \in \mathbb{R}^{h \times w \times c}$, would be adequate for constructing a higher-resolution latent $\bar{\mathbf{z}}_{T'} \in \mathbb{R}^{\bar{h} \times \bar{w} \times c}$. In pursuing this objective, as shown in the middle of Fig.\,\ref{fig:framework}, we reassemble pixels from the low-resolution latents $\{\bar{\mathbf{z}}_{T'}^i\}_{i=1}^{L_1}$ to form a higher-resolution latent, prior to carrying out specific detail refinement. This process can be expressed as:
\begin{equation}
\begin{aligned}
    (\bar{\mathbf{z}}_{T'})_{x,y,:} = (\bar{\mathbf{z}}^{ m \cdot w_s+n+1}_{T'})_{m + 1, n + 1, :}, 
                   \,\,\, m =(x -1) \% w_s , 
                   \,\,\, n = (y-1) \% h_s,
\end{aligned}
\end{equation}
%
where $h_s = \bar{h}/{h}$, $w_s = \bar{w}/{w}$, and the ``$\%$'' computes the remainder. 
We relocate the identical spatial pixels of $\{\bar{\mathbf{z}}_{T'}^i\}_{i=1}^{L_1}$ in adjacent regions of $\bar{\mathbf{z}}_{T'}$. For instance, the first pixels of low-resolution patches are placed in the pixels situated in the top-left corner of the higher-resolution image. The consequent higher-resolution latent $\bar{\mathbf{z}}_{T'}$ is thus featured with a complete integration of structural information.


Following the acquisition of the reassembled higher-resolution latent $\bar{\mathbf{z}}_{T'}$, as shown in the bottom of Fig.\,\ref{fig:framework}, we proceed to extract overlapping patches $\{\bar{\mathbf{z}}_{T'}^{i}\}_{i=1}^{L_2}$ using a shifted window, following MultiDiffusion~\cite{bar2023multidiffusion} in Sec.\,\ref{sec:backgrounds}. As we advance through time steps from $t = T' \rightarrow 1$, we continue patch-wise denoising with Eq.\,(\ref{eq:denoising-individual}), averaging overlapping regions. This stage refines specific details in the high-resolution latent structure, following principles in Fig.\,\ref{fig:visualization}.

\section{Experimentation}\label{sec:experimentation}

\subsection{Experimental Setting}


We conduct experiments using the pretrained SDXL~\cite{podell2023sdxl} where $T=50$ defaultly to show CutDiffusion's simplicity, efficiency, cheapness, and performance. We evaluate inference time and GPU memory consumption on a single 3090 GPU.
Given CutDiffusion's tuning-free nature, we compare it with SDXL-DI~\cite{podell2023sdxl}, Attn-SF~\cite{jin2023logn}, ScaleCrafter~\cite{he2023scalecrafter}, MultiDiffusion~\cite{bar2023multidiffusion}, and DemoFusion~\cite{du2023demofusion}. 

We employ three metrics for quantitative evaluation: FID (Frechet Inception Distance)~\cite{heusel2017fid}, IS (Inception Score)~\cite{salimans2016is}, and CLIP Score~\cite{radford2021clip-score}. $\text{FID}_r$ and $\text{IS}_r$ compare high-resolution generated images to real images but require resizing to $299^2$. Following~\cite{du2023demofusion,chai2022Any-resolution-training}, we crop and resize 10 local patches at $1\times$ resolution to compute $\text{FID}_c$ and $\text{IS}_c$. CLIP score evaluates cosine similarity between image embeddings and text prompts. We use 10,000 Laion-5B~\cite{schuhmann2022laion-5b} images as real images and 1,000 text prompts for CutDiffusion to generate high-resolution images.

\subsection{Method Comparison}

In this subsection, we show \textit{almighty capabilities} of our CutDiffusion method \emph{w.r.t.} method simplicity, inference time, GPU cost and generation performance.

\subsubsection{Method Simplicity.}

It is important to note that our CutDiffusion renders MultiDiffusion~\cite{bar2023multidiffusion} a special case when $T' = T$, under which circumstances no comprehensive structure denoising is applied. We merely modify the sub-patch approach as random Gaussian sampling, incorporating additional pixel interaction and relocation operations. Consequently, CutDiffusion retains the simplicity of MultiDiffusion while simultaneously reducing the number of patches required in the initial stage. Compared to more complex third-party engagements, such as fine-tuning~\cite{xie2023diff-fit,guo2024make-a-cheap-scaling}, low-resolution image generation~\cite{du2023demofusion}, and progressive upscaling~\cite{du2023demofusion},  our CutDiffusion method provides a more streamlined and straightforward process for generating higher-resolution images.

\subsubsection{Inference Time.}
The results in Table\,\ref{tab:inference_time} align with our conclusion in Table\,\ref{tab:cons-and-pros} that direct higher-resolution inference methods typically deliver faster speeds. An exception to this trend is ScaleCrafter~\cite{he2023scalecrafter}, whose use of larger kernels impacts inference speed, particularly in ultra-high-resolution settings. Despite adopting a patch-wise inference approach, our proposed CutDiffusion method achieves rapid inference speeds. These speeds are not only comparable to those achieved by direct inference methods~\cite{podell2023sdxl,jin2023logn} but are also significantly faster than those achieved by patch-wise counterparts~\cite{bar2023multidiffusion,du2023demofusion}. For instance, CutDiffusion is three times faster than DemoFusion~\cite{du2023demofusion}. %
CutDiffusion's speed advantage stems from its need for fewer inference patches during comprehensive denoising, unlike MultiDiffusion's many sub-patches. DemoFusion's prolonged inference time is due to its progressive upscaling strategy requiring more diffusion steps.

\begin{table}[!t]
\caption{A comparison of inference time among various training-free image generation extrapolation methods. The symbol ``$^*$'' denotes patch-wise denoising techniques.}
\vspace{-0.5em}
\label{tab:inference_time}
\centering
    \setlength\tabcolsep{8pt}
    \renewcommand{\arraystretch}{1.25}
    \resizebox{\textwidth}{!}{
        \begin{tabular}{c|cccccc}
        \toprule
\multirow{2}{*}{Resolution}     & \multicolumn{6}{c}{Method}          \\
\cline{2-7}
& SDXL-DL & Attn-SF & ScaleCrafter & MultiDiffusion$^*$ & DemoFusion$^*$ & CutDiffusion$^*$ \\

\midrule

1024 $\times$ 1024 (1$\times$)  
&  13s       &  -       &    -      &       -       &     -    &  - \\
\midrule

1024 $\times$ 2048 (2$\times$)  
& 28s     &  28s       &   31s       &  37s  &   82s   & 33s \\
\midrule

2048 $\times$ 1024 (2$\times$)  
& 28s      &  28s       &  31s       &  37s  & 82s  & 33s\\
\midrule

2048 $\times$ 2048 (4$\times$)  
&68s & 68s &  75s  &  113s & 207s  &  86s\\
\midrule

3072 $\times$ 3072 (9$\times$)  
&  216s  & 216s &  415s    &  314s   &  703s  &  224s \\


\bottomrule
\end{tabular}
}
\end{table}

\begin{table}[!t]

\caption{Minimal GPU memory requirements for executing existing image extrapolation methods. The symbol ``$^*$'' denotes patch-wise denoising techniques.}
\vspace{-0.5em}
\label{tab:gpu_cost}
\centering
    \setlength\tabcolsep{8pt}
    \renewcommand{\arraystretch}{1.25}
    \resizebox{\textwidth}{!}{
        \begin{tabular}{c|cccccc}
        \toprule
\multirow{2}{*}{Resolution}     & \multicolumn{6}{c}{Method}                                     \\
\cline{2-7}
                                & SDXL-DI          & Attn-SF       & ScaleCrafter   &  MultiDiffusion$^*$ & DemoFusion$^*$  & CutDiffusion$^*$ \\
                                \midrule
1024 $\times$ 1024 (1$\times$)  &  7.87G       & -             &  -             &     -           &   -         & -          \\
\midrule
2048 $\times$ 2048 (4$\times$)  &  9.60G       &  9.60G       &    9.76G      &   7.87G        &  7.87G     &  7.87G   \\
\midrule
3072 $\times$ 3072 (9$\times$)  & 12.52G       & 12.52G      &     17.48G        &  7.87G         &   7.87G    &  7.87G  \\
    \bottomrule
    \end{tabular}
}
\end{table}

\subsubsection{GPU Cost.}

Table\,\ref{tab:gpu_cost} shows minimal GPU memory needs for image extrapolation methods. Direct methods have faster inference but higher GPU demands, especially at increased resolutions~\cite{podell2023sdxl,jin2023logn,he2023scalecrafter}. Patch-based methods, like CutDiffusion, offer consistently lower peak memory usage, making them more accessible~\cite{bar2023multidiffusion,du2023demofusion}. Besides, CutDiffusion's reduced patch count in comprehensive structure denoising mitigates the long duration of GPU running~\cite{bar2023multidiffusion,du2023demofusion}.


\begin{figure}[!t]
    \centering
    \includegraphics[width=0.98\textwidth]{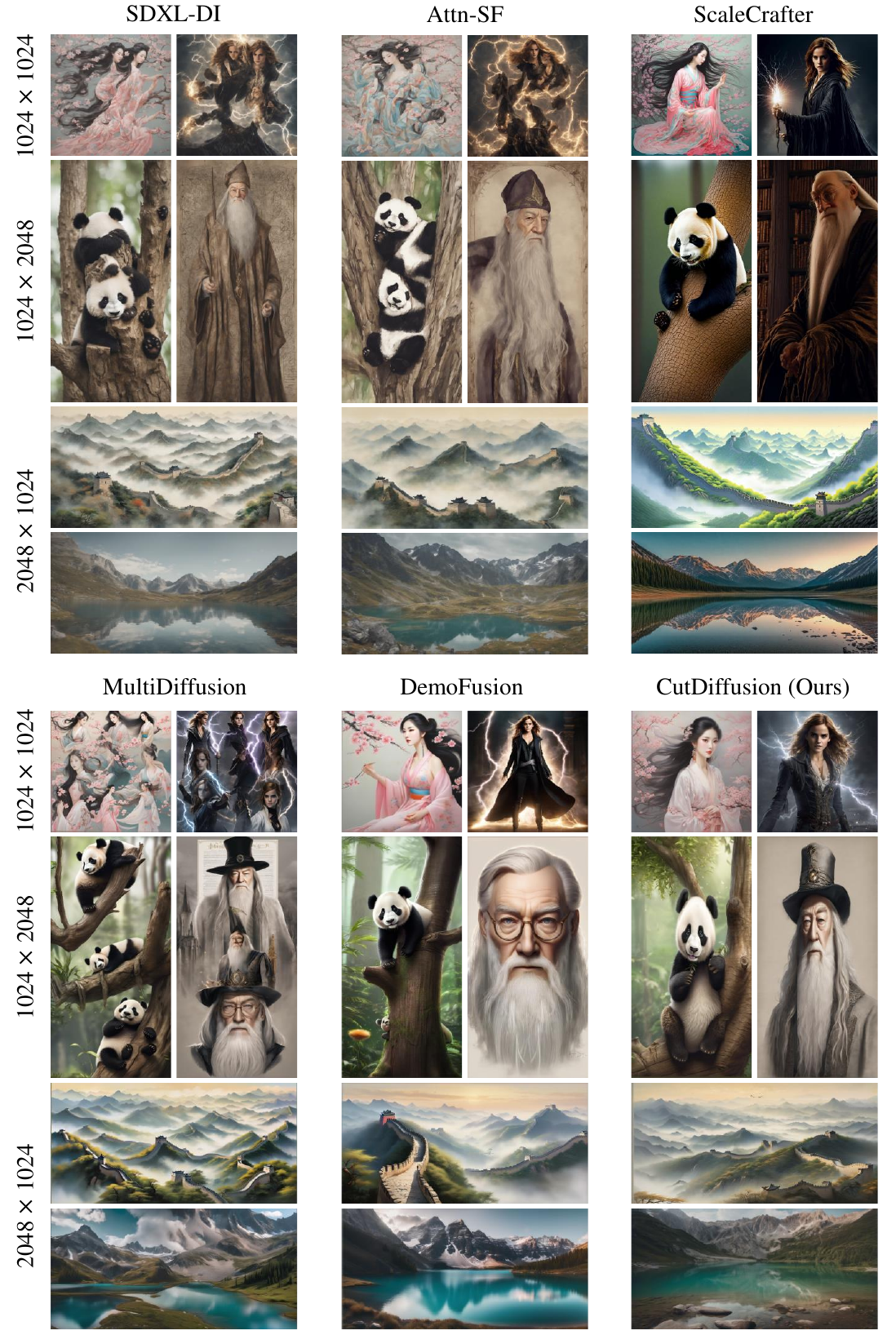}
    \caption{A comparison of higher-resolution generation. Best view with zooming in.}

    
    \label{fig:qualitative}
    \vspace{-1.0em}
\end{figure}

\subsubsection{Generation Performance.}
\label{sec:performance}
%
We further present visual results for qualitative comparison and metric performance for quantitative comparison, showcasing the strong generation capabilities of our proposed CutDiffusion.

\begin{table}[!t]
\caption{A comparison of quantitative metrics among various training-free image generation extrapolation methods. The best result is highlighted in \textbf{bold}, the second best in \underline{underline}, and patch-wise denoising methods are marked with $^*$.
}
\label{tab:quantitative comparisons}
\centering
    \setlength\tabcolsep{7pt}
    \renewcommand{\arraystretch}{1.25}
    \resizebox{\textwidth}{!}{
        \begin{tabular}{c|l|cccccc}
        \toprule
\multirow{2}{*}{Resolution}     & \multirow{2}{*}{Metrics} & \multicolumn{6}{c}{Method}   \\
\cline{3-8}

 &        & SDXL-DL     & Attn-SF    &ScaleCrafter        & MultiDiffusion*         & DemoFusion*   & CutDiffusion*        \\
\midrule

\multirow{5}{*}{1024 $\times$ 1024 (1$\times$)} 
& $\text{FID}_r\downarrow $ & 58.49        &  -       &   -             &  -            &  -     &  -   \\
& $\text{IS}_r\uparrow$     &  17.39       &  -       &  -              &  -            &  -     &  -   \\
& $\text{FID}_c\downarrow$  &  58.08       &   -      &   -             &  -            &  -     &  -   \\
& $\text{IS}_c\uparrow$     &  25.38       &  -       &   -             &  -            &  -     &  -  \\
& CLIP$\uparrow$            &  33.07       &  -       &   -             &   -           &   -    &  -   \\
\midrule

\multirow{5}{*}{1024 $\times$ 2048 (2$\times$)} 
& $\text{FID}_r\downarrow $ & 89.38       &  89.12                &  96.37    &   \underline{76.06}           &  78.05     &  \textbf{66.31}   \\
& $\text{IS}_r\uparrow$     & 12.52       &  \underline{12.75}              &  12.34    &  11.69          &  10.49    & \textbf{16.68}   \\
& $\text{FID}_c\downarrow$  & 59.98    &   59.23               &  76.69      &    \textbf{41.32}          &  47.64     &   \underline{44.67}   \\
& $\text{IS}_c\uparrow$     &  18.46   &  18.74                &  15.37     &   \underline{20.66}           &   17.79     &  \textbf{23.95 } \\
& CLIP$\uparrow$            &  29.60       &  29.72                  &   27.67    &  \textbf{ 32.75 }        &   30.55    &   \underline{30.70}   \\
\midrule

\multirow{5}{*}{2048 $\times$ 1024 (2$\times$)} 
& $\text{FID}_r\downarrow $ & 93.01       &  92.79                  &  89.27     &   \underline{66.56}         &  73.56     &  \textbf{62.64}   \\
& $\text{IS}_r\uparrow$     & 11.47       &  11.66                  &  11.19     &  \underline{13.79}        &  11.91     &  \textbf{17.39}   \\
& $\text{FID}_c\downarrow$  & 64.51       &   64.12                 &  67.79        &   \textbf{39.33}      &  52.86     &  \underline{44.73}   \\
& $\text{IS}_c\uparrow$     & 18.90       &  18.97                 &  15.34          &   \underline{23.90}     &  18.60     &  \textbf{24.81}  \\
& CLIP$\uparrow$            & 28.98       &  29.03                    &   29.64   &   \textbf{32.18}        &   30.36    &  \underline{30.42}   \\
\midrule

\multirow{5}{*}{2048 $\times$ 2048 (4$\times$)}  
& $\text{FID}_r\downarrow $ & 124.40     &  124.15          & 99.47   &   81.46   &  \textbf{60.46}     &  \underline{72.57}  \\
& $\text{IS}_r\uparrow$     & 11.05      & 11.15           & 12.52   &  12.43     &  \textbf{16.45}     &  \underline{16.02}  \\
& $\text{FID}_c\downarrow$  & 88.33      &  88.59           & 74.64  & \underline{44.80}   &  \textbf{38.55}     &  47.60  \\
& $\text{IS}_c\uparrow$     & 14.64      & 14.81            & 15.42  &  20.99     &  \textbf{24.17}      &  \underline{23.13}   \\
& CLIP$\uparrow$            &  28.11     & 28.12             & 28.82 &   \underline{31.82}    &   \textbf{32.21}    &  31.13   \\
\midrule

\multirow{5}{*}{3072 $\times$ 3072 (9$\times$)}  
& $\text{FID}_r\downarrow $ &  170.61       &  170.62                   &     131.42   &   101.17      &   \textbf{63.03}    &  \underline{101.14}  \\
& $\text{IS}_r\uparrow$     &  7.83         &   7.93                          &   9.62     &    8.83           &  \textbf{16.41}      &    \underline{11.94} \\
& $\text{FID}_c\downarrow$  &  112.51       &   112.46               &   105.79      &     \underline{51.95}       &  \textbf{48.45}      &   77.44   \\
& $\text{IS}_c\uparrow$     &  12.59        &  12.52                   &    11.91   &     \underline{17.74}       &  \textbf{20.42}      &   14.54  \\
& CLIP$\uparrow$            &  24.53        &   24.56                &  27.22         &    \underline{29.49}      &   \textbf{32.25}     &   28.25 \\
    \bottomrule
    \end{tabular}
}
\end{table}

\textit{Qualitative Comparison. }
We present qualitative comparisons under different aspect ratios (1:1, 1:2, and 2:1) and extrapolation ratios ($2\times$, $4\times$, and $9\times$). In Fig.\,\ref{fig:qualitative},\footnote{Specific prompts and additional results are provided {\color{blue}in the Appendix.}} SDXL-DI~\cite{podell2023sdxl} exhibits severe repetition and structural distortion. Despite adjusting the attention scaling factor, Attn-SF~\cite{jin2023logn} struggles with the same issues. Although ScaleCrafter~\cite{he2023scalecrafter} generates non-repetitive images, local details like human hands and panda feet lack fidelity. MultiDiffusion~\cite{bar2023multidiffusion} suffers from repetitive and inconsistent content due to insufficient global structural information. DemoFusion~\cite{du2023demofusion} occasionally produces repeated small objects, such as the small panda beneath the tree trunk. Owing to the collaborative two-stage generation process, CutDiffusion results in high-quality, higher-resolution images.

\textit{Quantitative Comparison.}
%
%
Table\,\ref{tab:quantitative comparisons} demonstrates that CutDiffusion outperforms other image extrapolation methods, achieving the best $\text{FID}_r$ scores at resolutions of $1024\times2048$ and $2048\times1024$. Although CutDiffusion's $\text{FID}_c$ and CLIP score are marginally lower than MultiDiffusion's~\cite{bar2023multidiffusion}, it is important to note that $\text{FID}_c$ indicates detail generation ability. MultiDiffusion generates local details throughout the entire process, while CutDiffusion does so only in the second stage, resulting in slightly inferior detail. Nevertheless, in Fig.\,\ref{fig:qualitative}, MultiDiffusion exhibits significant object repetitions despite its higher CLIP and $\text{FID}_c$. Given both quantitative and qualitative results, CutDiffusion surpasses MultiDiffusion. 

\begin{figure}[!t]
    \centering
    \includegraphics[width=0.95\textwidth]{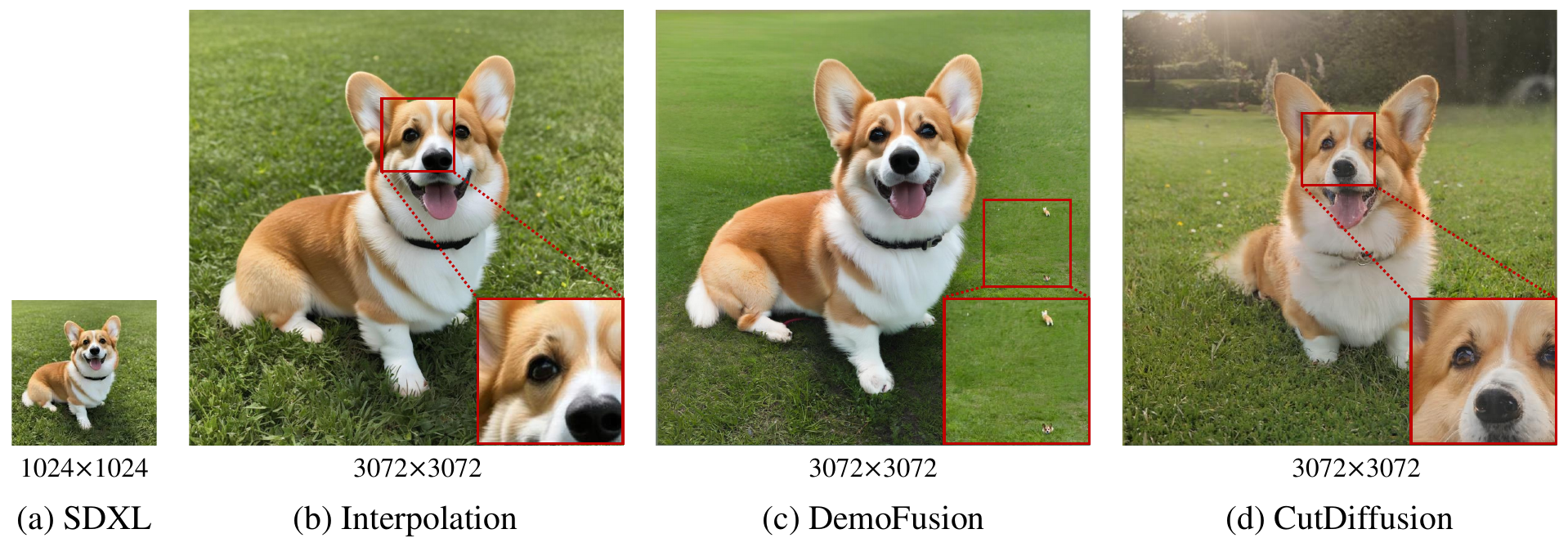}
    \caption{Visual results of ``A cute corgi on the lawn.'' Best view with zooming in. 
    }
    \label{fig:explain_fid}
    \vspace{-1.0em}
\end{figure}

Progressing to the resolutions of $2048\times 2048$ and $3072\times3072$, CutDiffusion seems to achieve the second-best. A closer look at Fig.\,\ref{fig:explain_fid} reveals that DemoFusion's skip connection generates a $3072\times3072$ image with the same content as a pre-trained $1024\times1024$ image~\cite{podell2023sdxl}. This implies that \textit{DemoFusion's performance advantage is due to the low-resolution rather than its inherent capabilities}. To confirm this, we interpolate images from $1024 \times 1024$ to $3096 \times 3096$ resolution, obtaining the best $\text{FID}_r$ (58.50) and $\text{IS}_r$ (17.38). Thus, we conclude that \textit{mimicking the low-resolution distribution yields a higher FID score from the pre-trained SDXL model}. However, as shown by Fig.\,\ref{fig:qualitative} and Fig.\,\ref{fig:explain_fid}, this approach leads to repetition issues, and even lacks high-resolution details for interpolation. Excluding unfair comparison with DemoFusion, our CutDiffusion attains the best.

\subsection{Ablation Study}\label{sec:ablation}
We shift our focus to ablate CutDiffusion, simple for its single hyper-parameter $T'$. We also ablate the pixel interaction. All ablations are conducted using prompt $\mathcal{P}$ = ``A photo of cat.'' with $2048\times2048$ resolution under different random seeds.

\subsubsection{Ablations on Hyper-Parameter.}
\begin{figure}[!b]
    \centering
    \includegraphics[width=0.95\textwidth]{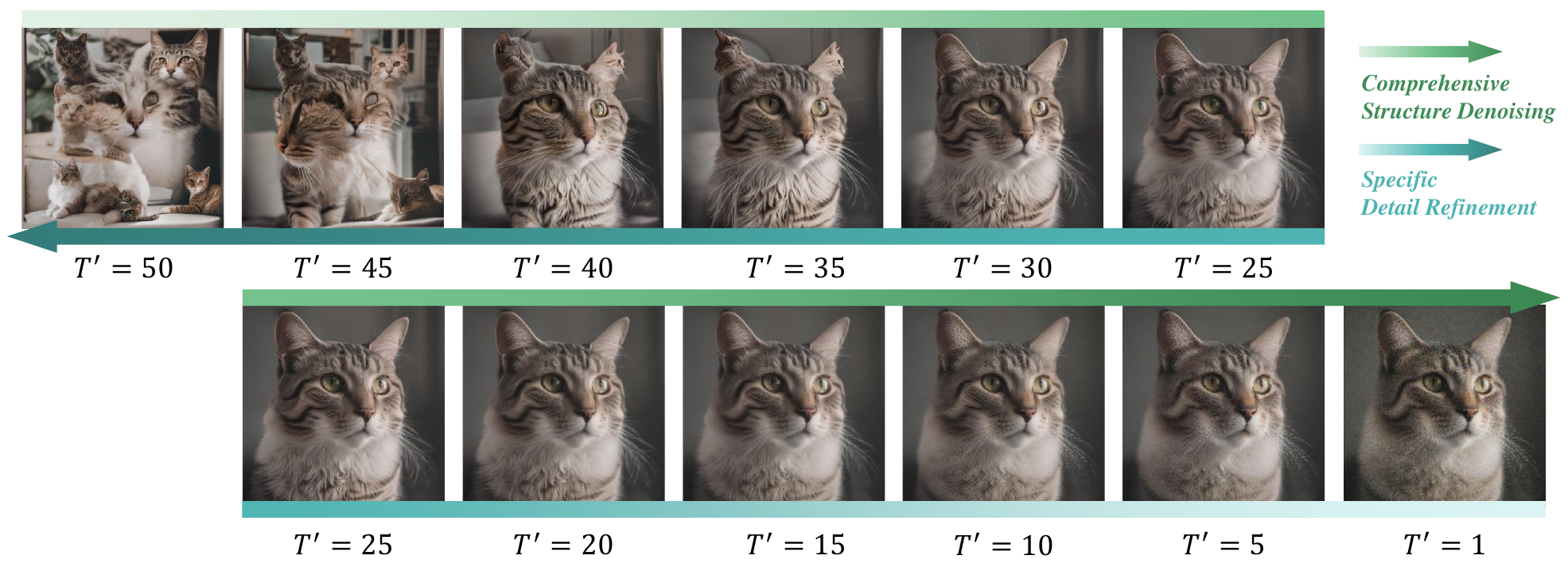}
    \caption{Higher-resolution images for different $T'$ values. Best view with zooming in.}
    \label{fig:ablation1}
    \vspace{-1.0em}
\end{figure}

CutDiffusion cuts denoising process into two stages: comprehensive structure denoising and specific detail refinement, regulated by hyper-parameter $T'$. A higher $T'$ means more steps to detail refinement and less to structure denoising, and vice versa. Fig.\,\ref{fig:ablation1} shows $T' = 50$ guides CutDiffusion to resemble MultiDiffusion~\cite{bar2023multidiffusion}, causing repeated objects; $T' = 1$ emphasizes structure denoising, leading to a clear structure but vague details. Visual assessment indicates that $T' = T/2 = 25$ strikes the optimal balance.

\begin{figure}[!t]
    \centering
    \includegraphics[width=0.95\textwidth]{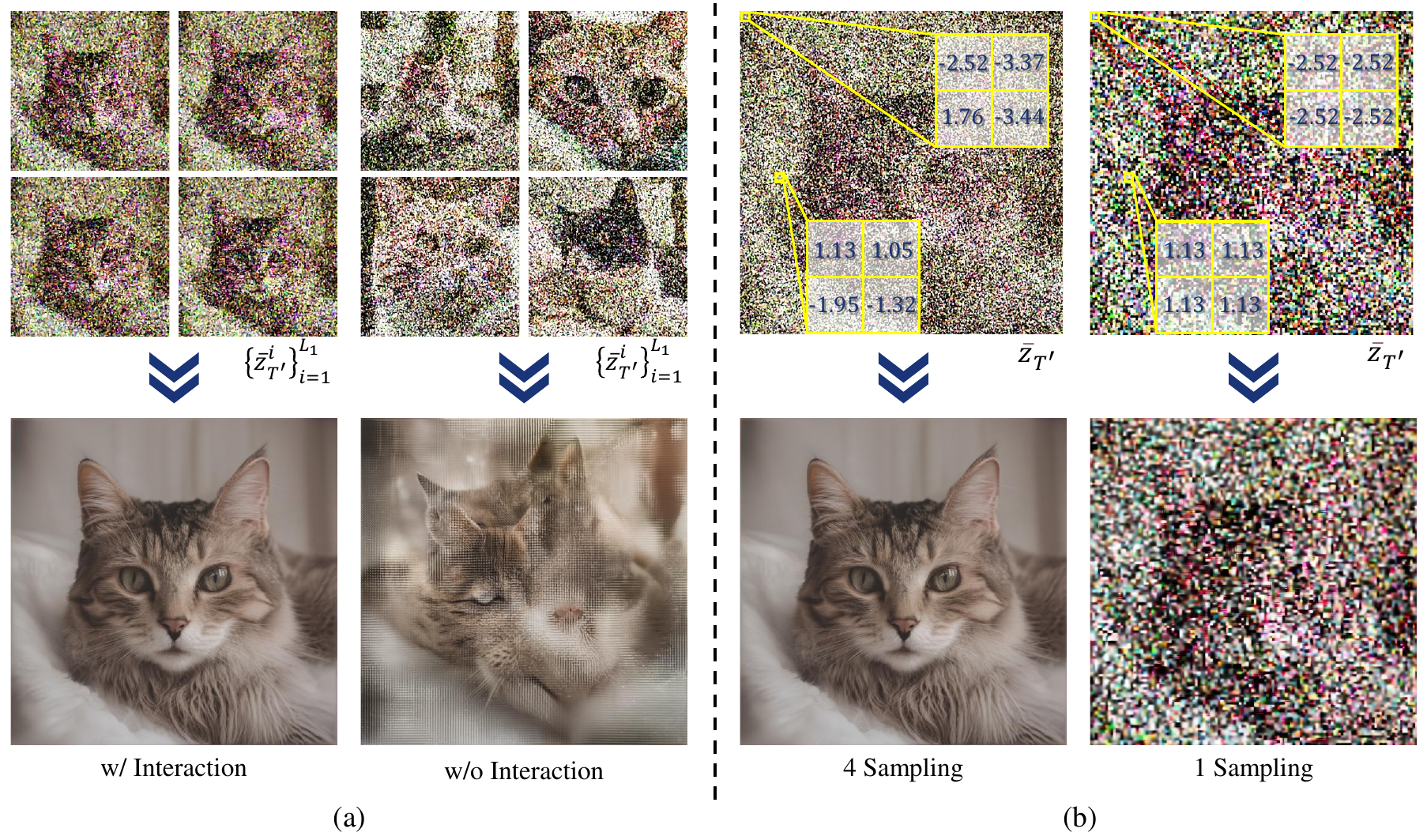}
    \caption{Results of (a) with/without pixel interaction and (b) 4 sampling \emph{v.s.}  1 sampling. }
    \label{fig:ablation_modules}
    \vspace{-1.0em}
\end{figure}

\subsubsection{Ablations on Pixel Interaction.}
As outlined in Sec.\,\ref{sec:denoising}, our pixel interaction guarantees images not only pertain to the same subject but also share similar content. Fig.\,\ref{fig:ablation_modules}(a) substantiates this assertion. The pixel interaction ensures the sub-patches depict similar cats, culminating in a high-quality, high-resolution image. Conversely, the absence of pixel interaction results in sub-patches featuring distinct cats, thereby compromising the generation of high-resolution images.

Another potential way to ensure consistent content involves sampling a single Gaussian noise instead of $L_1 = 4$, creating four copies in the step $T'$, and executing pixel relocation for a higher-resolution latent $\bar{\mathbf{z}}_{T'}$. However, as illustrated in Fig.\,\ref{fig:ablation_modules}(b), this method fails due to the copy operation. The copy operation generates identical pixels in neighboring areas, disrupting the Gaussian distribution and causing irregular denoising during specific detail refinement stages.

\section{Limitation and Future work}
Our CutDiffusion method has limitations: (1) Similar to other methods, the quality of high-resolution images generated by CutDiffusion is somewhat dependent on the pretrained diffusion model. (2) The second stage of CutDiffusion still necessitates as many patches as existing methods, hindering further acceleration.

Future work may investigate higher-resolution generation in a non-overlapping patch-wise fashion, with the potential to enhance inference speeds further.

\section{Conclusion}
We have proposed a mighty diffusion extrapolation method, CutDiffusion, which streamlines and accelerates the process while enhancing performance and affordability. CutDiffusion cuts the standard patch diffusion process into an initial comprehensive structure denoising and a subsequent specific detail refinement. It only modifies the sub-patch sampling approach, integrating pixel interaction and relocation in the respective stages. Unlike existing methods that excel in specific areas, CutDiffusion highlights in all perspectives of simple method construction, fast inference speed, cheap GPU cost and strong generation performance.

\par\vfill\par


%
%
\bibliographystyle{splncs04}
\bibliography{main}

\appendix

\section{More Qualitative Results}
In Fig.\,\ref{fig:qualitative-1}, Fig.\,\ref{fig:qualitative-2} and Fig.\,\ref{fig:qualitative-3}, we present an extensive collection of visually compelling results generated by our innovative CutDiffusion method at various higher-resolution levels, showcasing the effectiveness and versatility of our approach.

\section{Prompt List}

We provide specific prompts we use to generate Fig.\,\ref{fig:qualitative} in the main paper.

\begin{enumerate}
    \item A painting of a beautiful graceful woman with long hair, a fine art painting, by Qiu Ying, no gradients, flowing sakura silk, beautiful oil painting.
    \item Emma Watson as a powerful mysterious sorceress, casting lightning magic, detailed clothing.
    \item A cute panda on a tree trunk. 
    \item A Portrait of Albus Dumbledore. 
    \item The Great Wall of China winding through mist-covered mountains, captured in the delicate brushwork and harmonious colors of a traditional Chinese landscape painting.
    \item RAW photo of a mountain lake landscape, clean water, 8k, UHD.
\end{enumerate}

\begin{figure}[!t]
    \centering
    \includegraphics[width=0.86\textwidth]{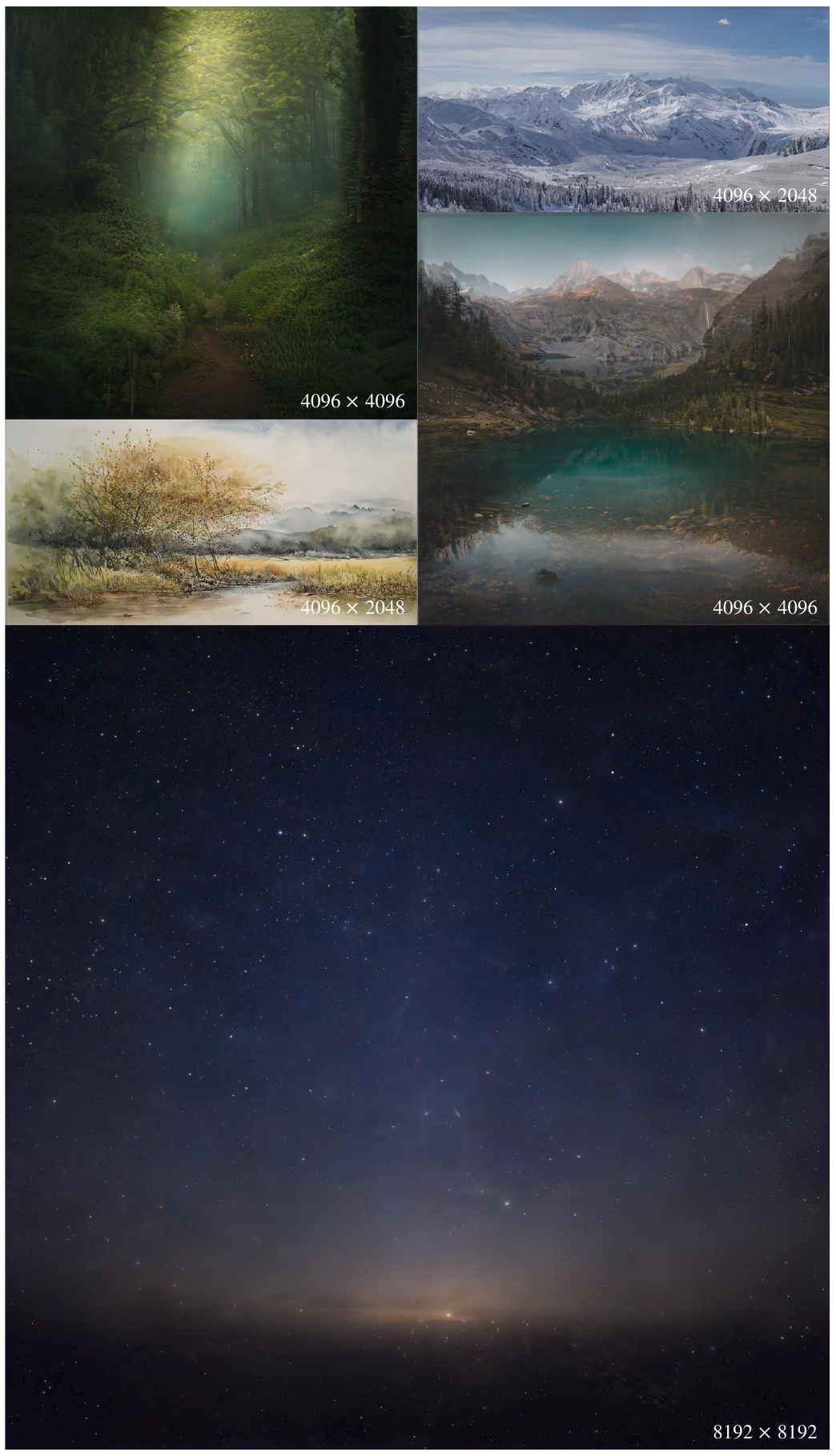}
    \caption{Visual results of our CutDiffusion on SDXL (1024$^2$) with various higher-resolution resolutions. Best viewed with zooming in.}

    \label{fig:qualitative-1}
\end{figure}

We provide specific prompts we use to generate Fig.\,\ref{fig:qualitative-1} in the main paper.

\begin{enumerate}
    \item A fantasy forest.
    \item Watercolor Painting, Handmade, Gift, Wall Hanging, Home Decor, Landscape, Nature, Size/ Height 12.9 Inches X Width 10 Inches.
    \item Snow-capped mountain.
    \item A picturesque mountain scene with a clear lake reflecting the surrounding peaks.
    \item The vast starry sky.
\end{enumerate}

\begin{figure}[!t]
    \centering
    \includegraphics[width=0.98\textwidth]{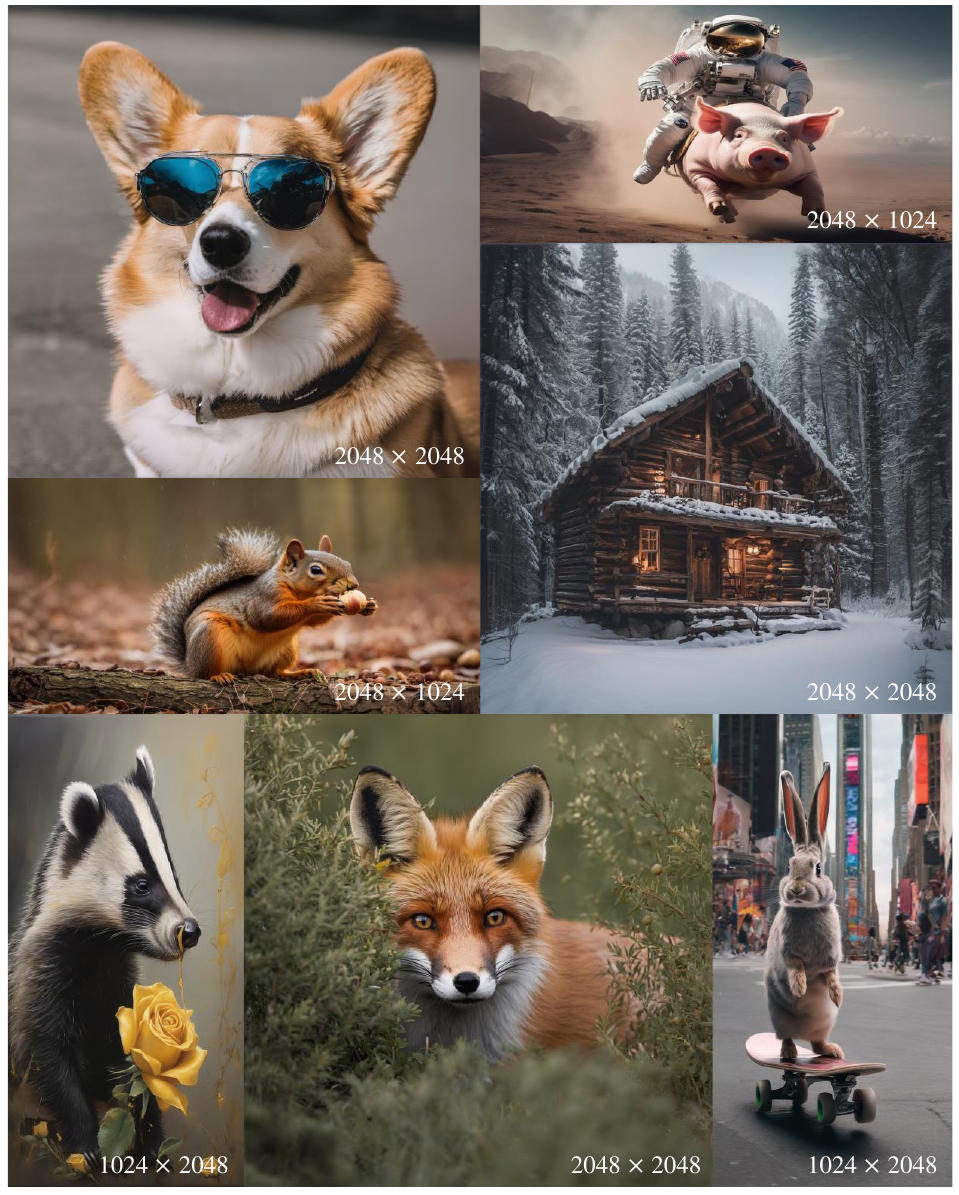}
    \caption{Visual results of our CutDiffusion on SDXL (1024$^2$) with various higher-resolution resolutions. Best viewed with zooming in.}
    \label{fig:qualitative-2}
\end{figure}

We provide specific prompts we use to generate Fig.\,\ref{fig:qualitative-2} in the main paper.

\begin{enumerate}
    \item A corgi wearing cool sunglasses. 
    \item An astronaut riding a pig, highly realistic dslr photo, cinematic shot.
    \item A squirrel eating an acorn in a fores.
    \item A rustic wooden cabin nestled in a snowy forest.
    \item A young badger delicately sniffing a yellow rose, richly textured oil painting.
    \item A fox peeking out from behind a bush.
    \item A rabbit is skateboarding in Time Square.
\end{enumerate}

\begin{figure}[!t]
    \centering
    \includegraphics[width=0.98\textwidth]{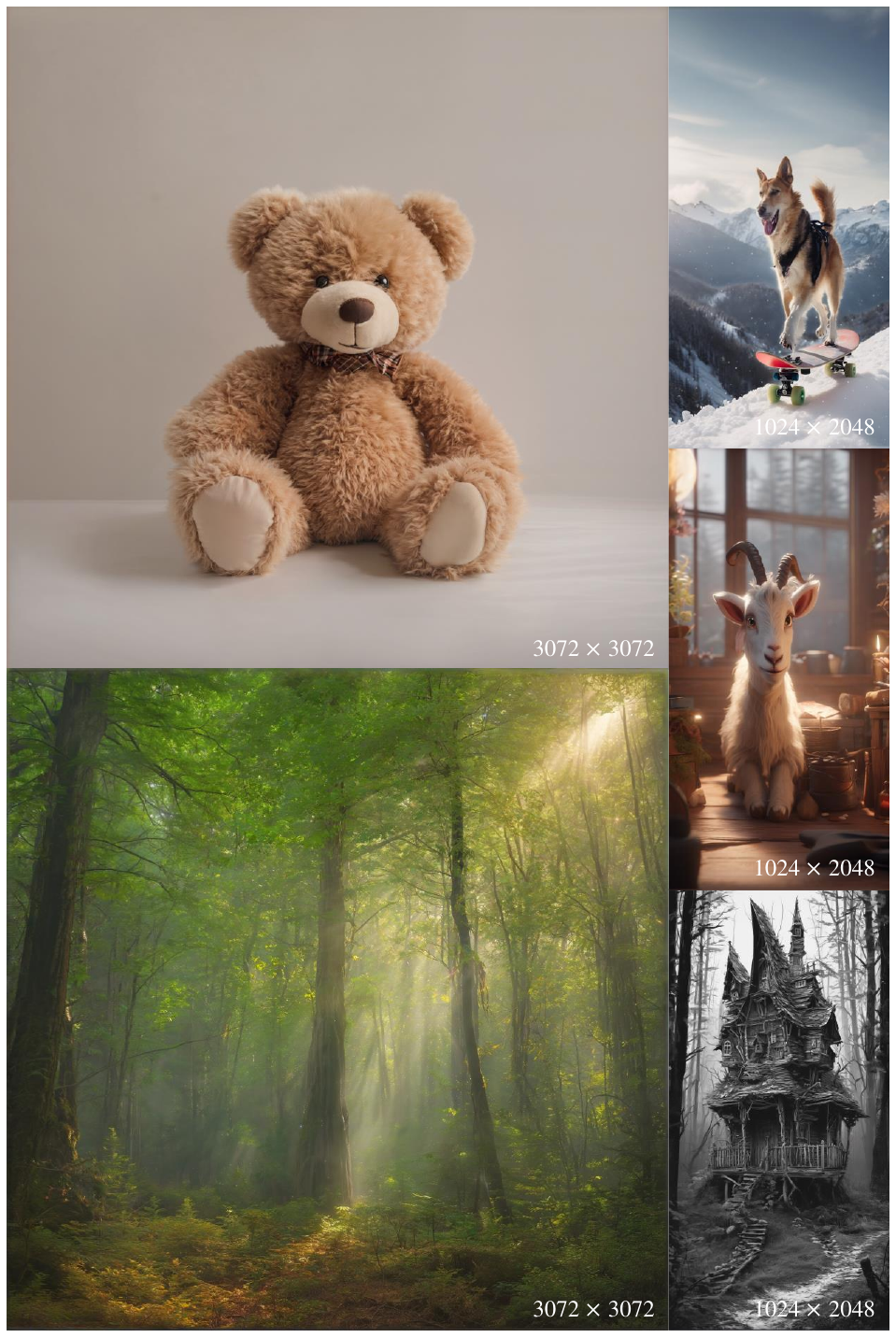}
    \caption{Visual results of our CutDiffusion on SDXL (1024$^2$) with various higher-resolution resolutions. Best viewed with zooming in.}
    \label{fig:qualitative-3}
\end{figure}

We provide specific prompts we use to generate Fig.\,\ref{fig:qualitative-3} in the main paper.

\begin{enumerate}
    \item A cute teddy bear in front of a plain white wall, warm and brown fur, soft and fluffy.
    \item Primitive forest, towering trees, sunlight falling, vivid colors.
    \item A Dog is skateboarding on the snowy mountain.
    \item Cute adorable little goat, unreal engine, cozy interior lighting, art station, detailed’ digital painting, cinematic, octane rendering.
    \item Monster Baba yaga house with in a forest, dark horror style, black and white.
\end{enumerate}

\end{document}